\documentclass[letterpaper, 10 pt, conference]{ieeeconf}  

\IEEEoverridecommandlockouts                              
\usepackage{graphicx}

\title{\LARGE \bf
SceneEDNet: A Deep Learning Approach for Scene Flow Estimation
}

\author{Ravi Kumar Thakur and Snehasis Mukherjee
\thanks{Ravi Kumar Thakur is with Computer Vision Group, Indian Institute of Information Technology, Sri City, Chittoor
        {\tt\small ravikumar.t@iiits.in}}%
\thanks{Snehasis Mukherjee is with Computer Vision Group, Indian Institute of Information Technology, Sri City, Chittoor
        {\tt\small snehasis.mukherjee@iiits.in}}%
}

\begin{document}

\maketitle
\thispagestyle{empty}
\pagestyle{empty}

\begin{abstract}
Estimating scene flow in RGB-D videos is attracting much interest of the computer vision researchers, due to its potential applications in robotics. The state-of-the-art techniques for scene flow estimation, typically rely on the knowledge of scene structure of the frame and the correspondence between frames. However, with the increasing amount of RGB-D data captured from sophisticated sensors like Microsoft Kinect, and the recent advances in the area of sophisticated deep learning techniques, introduction of an efficient deep learning technique for scene flow estimation, is becoming important. This paper introduces a first effort to apply a deep learning method for direct estimation of scene flow by presenting a fully convolutional neural network with an encoder-decoder (ED) architecture. The proposed network SceneEDNet involves estimation of three dimensional motion vectors of all the scene points from sequence of stereo images. The training for direct estimation of scene flow is done using consecutive pairs of stereo images and corresponding scene flow ground truth. The proposed architecture is applied on a huge dataset and provides meaningful results.
\end{abstract}

\section{Introduction}
Motion estimation in videos, has been an active area of research in computer vision during the last few decades, due to its potential applications in robot navigation, healthcare, surveillance, elderly and child monitoring, human-robot interaction, structure from motion, three-dimensional reconstruction and autonomous driving and many more, by understanding the appearance of objects in motion \cite{Yan2016SceneFE}. The first step in motion estimation is to compute displacement vectors between different points. Scene flow is a three dimensional displacement vector of scene points moving between different time steps. Scene flow can be considered as the three dimensional form of optical flow. Scene flow provides the motion vectors of the scene points in three dimensional space: along x, y and z axes. This can be used to obtain full information and geometry of the scene objects. A sufficient knowledge of scene flow is important for scene understanding during motion in RGB-D videos.
   \begin{figure}[t]
      \includegraphics[width=\columnwidth,height=7cm]{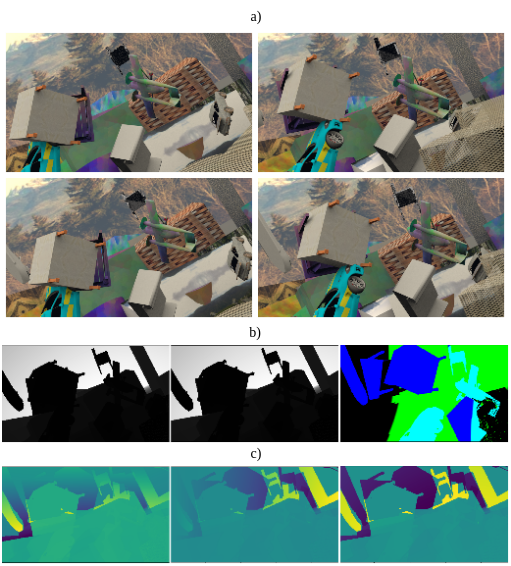}
      \caption{a) Shows pairs of stereo images from FlyingThings3D dataset b) Corresponding disparities and left optical flow c) Reconstructed scene flow along x-, y- and z-directions.}
      \label{figure1}
  \end{figure}

For a complete definition of scene flow, stereo image pairs at different time steps are required. It can also computed using RGB-D image pairs where the depth data is directly available. The scene flow can be constructed using optical flow and disparity obtained from stereo pairs. Figure \ref{figure1} shows stereo image pairs with its ground truth disparities, optical flow and scene flow along the three orthogonal directions. Hence, the accuracy of computed scene flow depends on accurate estimation of optical flow and disparity. A small error in sub-components optical flow and disparity will produce very large error in corresponding scene flow. Given the camera parameters and consecutive pairs of stereo images, problem of scene flow estimation is to compute three dimensional motion field of all the scene points along the three orthogonal directions. 
 
There are various challenges involved in scene flow estimation process. In case of occlusion, the obstruction can lead to inaccurate information of the scene points. If there are large displacements between the two consecutive frames then accuracy of estimation is less. Scene flow estimation also becomes difficult in outdoor scenes where variations in illumination are present. Insufficient texture also poses challenge to the estimation process, as it becomes difficult to calculate the motion when pixels look alike.

Several classical or non-learning based techniques are developed specifically to tackle some of these challenges. Most of these state-of-the-art methods can be considered as an extension of the optical flow estimation process to 3D. The classical methods of scene flow estimation often take time ranging from several seconds to minutes. Thus, they are not suitable for real-time applications. With the recent advancements of sophisticated techniques to handle large amount of data, a deep learning based technique is necessary to estimate scene flow. Unfortunately, a very few efforts have been found in the literature to estimate scene flow by deep learning technique \cite{mayer2016large}.

We propose a deep learning approach for end-to-end learning of scene flow. The proposed technique take the advantage of availability of a large dataset of stereo images with corresponding ground truth optical flow and disparity. Though the training time for deep learning methods on large dataset is much higher, unlike classical techniques it is faster during runtime. Since the network is trained over large dataset, it learns the cases where assumptions such as brightness constancy and insufficient texture are violated. There are some works reported in estimation of optical flow and disparity using neural networks. However, reconstructing scene flow from these two components will be inaccurate depending upon the lack of accuracy of the optical flow and disparity estimation. Unlike classification, scene flow estimation through neural network is structured prediction problem where labels are provided for every pixel. As encoder decoder architectures have shown significant success in many reconstruction problems (such as 3D reconstruction \cite{3d}), this paper presents a fully convolutional encoder decoder architecture to estimate dense scene flow.

The contributions of this paper can be considered as two folds. First, we introduce an encoder decoder architecture for direct scene flow estimation (without going for estimation of optical flow and disparity) from a huge dataset. Second, we have annotated the FlyingThings3D dataset \cite{mayer2016large} by providing the ground truth scene flow along the x, y and z directions for each video, for training a deep learning architecture.

\section{Related Works}
In this section we first discuss a survey on scene flow estimation methods available in the literature. This discussion will be followed by another discussion on the recent efforts for structure prediction using deep learning models.

\subsection{Scene Flow Estimation}
The first work on scene flow was reported by Vedula \textit{et al.} \cite{vedula1999three}. The scene flow was computed from optical flow for three different scenarios based on knowledge of scene structure and correspondences. The method was developed for lambertian surface. A piecewise rigid model was introduced by Vogel \textit{et al.} \cite{vogel2013piecewise}, where the scene was described as rigidly moving segments. Then, the estimation of scene flow was done by performing  joint estimation of segmentation of planar and rigid regions in the scene. Another approach of estimating scene flow by dividing scene into planar superpixel was introduced by Menze \textit{et al.} \cite{menze2015object}. The scene flow was estimated by optimizing a conditional random field. This work also introduced a challenging dataset for scene flow. Another method for scene flow estimation using stereo sequences was proposed by Huguet et al. \cite{huguet2007variational}. The optical flow from both the cameras and dense stereo matching was used for scene flow estimation in \cite{huguet2007variational}. The use of stereo cameras for scene flow estimation is a classical technique. Since the advent of depth cameras, several researchers have started utilizing RGB-D images for scene flow estimation. Quiroga \textit{et al.} used twisted motion model on RGB-D scenes \cite{quiroga2014dense}. The scene flow was then formulated by solving for energy function consisting of data and smoothness term. Sun \textit{et al.} computed scene flow by layerwise modeling the RGB-D images \cite{sun2015layered}. The motion  flow for each layer was estimated separately.

Hadfield \textit{et al.} introduced a particle filter based method for scene flow \cite{hadfield2014scene}. Scene flow estimation by a seeding algorithm was introduced in \cite{vcech2011scene}, where optical flow and disparity were computed simultaneously from stereo images. Basha \textit{et al.} introduced an energy function consisting of multi-view information,  where both depth and scene flow were estimated simultaneously \cite{basha2013multi}. To deal with occlusions and non-rigid deformations of objects in the video during motion, Golyanik \textit{et al.} \cite{golyanik2016nrsfm} proposed scene flow estimation from monocular images. This approach does not depend on assumptions such as camera details. All the approaches for scene flow estimation discussed so far, are tested on datasets of limited size. The first attempt of estimating scene flow on a large dataset is found in \cite{golyanik2017multiframe}. However, very few efforts \cite{mayer2016large} has been found in the literature to estimate scene flow with deep learned features.

\subsection{Per-Pixel Structure Prediction using CNN}
Convolutional neural network (CNN) are well known for their performance on image classification problem. The ability of CNN to extract highly abstract features have shown good results on various problems in the field of computer vision. Recently, efforts have been made to use CNNs in structure prediction tasks such as image segmentation, disparity estimation and flow estimation. In structured prediction problems every pixel has a corresponding label. Long \textit{et al.} proposed a fully convolutional network for semantic segmentation \cite{long2015fully}. In this work classification networks such as AlexNet, VGGNet were adapted as fully convolutional network. The network was trained to perform semantic segmentation. A CNN was trained to perform stereo matching by Zbontar \textit{et al.} \cite{zbontar2016stereo}. The network was trained on known set of images and disparity. This was followed by series of post processing steps such as cost aggregation and semi-global matching for refinement and accuracy. The final disparity map was obtained by applying sub-pixel enhancement and bilateral filter. A work on depth prediction using single image was done by Eigen \textit{et al.} \cite{eigen2014depth}. They trained two parallel CNNs on ground truth depth, making use of both local and global information. The first network predicts depth using global view of the scene, and the second network is trained to perform local refinement.

The introduction of a large dataset by Mayer \textit{et al.} \cite{mayer2016large} encouraged researchers to work on optical flow and disparity estimation using deep learning. The dataset is artificially rendered. It provides optical flow and disparity ground truth. They also introduced SceneFlowNet as combination of FlowNet and DispNet for estimation of scene flow. It was preceded by FlowNet by Fisher \textit{et al.} \cite{fischer2015flownet}. They trained a fully CNN for optical flow estimation on artificially generated FlyingChairs dataset. Two different networks based on input modes were introduced. This work was followed by FlowNet 2.0 by \cite{ilg2016flownet} which introduced dataset scheduling for improving the accuracy reported by FlowNet. The Flownet was the first attempt to estimate the optical flow in videos using CNN, which encouraged researchers to estimate optical flow using CNN. Ranjan \textit{et al.} introduced spatial pyramid network which has very small number of training parameters compared to Flownet \cite{ranjan2017optical}. The spatial pyramid here is used to deal with large motions. The convolutional layers then update optical flow at each layer. 

Motion field flow prediction in videos using CNNs, still remains an unsolved problem due to the complexity inherent into the videos for structure prediction. None of the deep learning based approaches for optical flow could produce better results compared to the state-of.the-art handcrafted techniques. Hence, mere extension of the optical flow prediction techniques to estimate scene flow, does not work well. However, the recent works in per-pixel prediction tasks using CNNs show the possibility that scene flow estimation can be performed by training a deep network on large dataset. Motivated by the recent efforts on optical flow estimation, we propose a technique to directly estimate the scene flow in x, y and z directions from the RGB-D videos, by introducing an encoder-decoder architecture. We apply the proposed method on FlyingThings3D \cite{mayer2016large} dataset. Although the results of the proposed approach cannot outperform the state-of-the-art handcrafted techniques, however, this first attempt of CNN architecture for  direct estimation of scene flow can be enhanced further to be used for dealing with large datasets.


\addtolength{\textheight}{-3cm}   

\section{Proposed Methodology}
In this section we describe the form of data used for the proposed method, followed by a detailed description of the proposed CNN architecture.

\subsection{Data Model}
\begin{figure}[!t]
      \centering
      \includegraphics[width=8cm]{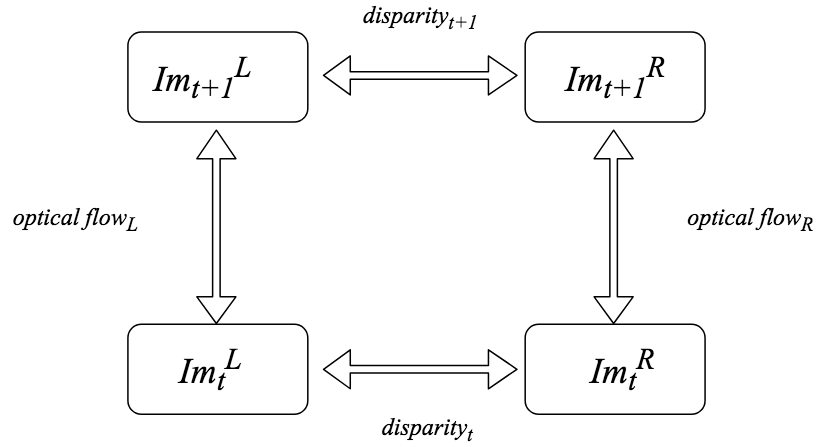}
      \caption{Relation between stereo image pairs. The images at  consecutive time steps give left and right optical flow. Whereas, the stereo pairs gives disparities.}
      \label{figure2}
  \end{figure}

For training the proposed network we use sequence of stereo images as input. The stereo images are rectified. To get a complete definition of scene flow, stereo image at $t$ and $t+dt$ is sufficient. The optical flow is given by images at different time steps. Whereas, disparity is given by images from left and right camera. The optical flow gives us the two dimensional motion information. For our work we have used optical flow images from left camera. The disparity comes from stereo pair from consecutive time steps. The disparity gives information about displacement along the $z$ direction. Given the camera baseline and focal length the depth can be obtained. Thus, a change in disparity gives us the change in depth data. Figure \ref{figure2} shows data relation between input images. Using the camera extrinsic and intrinsic parameters, the ground truth scene flow can be  constructed from optical flow and change in depth. However, It should be noted that a small error in optical flow or disparity can result in a large error in the scene flow. Hence, we train the proposed CNN architecture with the ground truth images of the scene flow along the three orthogonal directions. The ground truths are obtained from the dataset by proper annotation process. We describe the annotation process in Dataset section.

\subsection{Network Architecture}
\begin{figure*}[thpb]
      \centering
      \includegraphics[width=\textwidth,height=6cm]{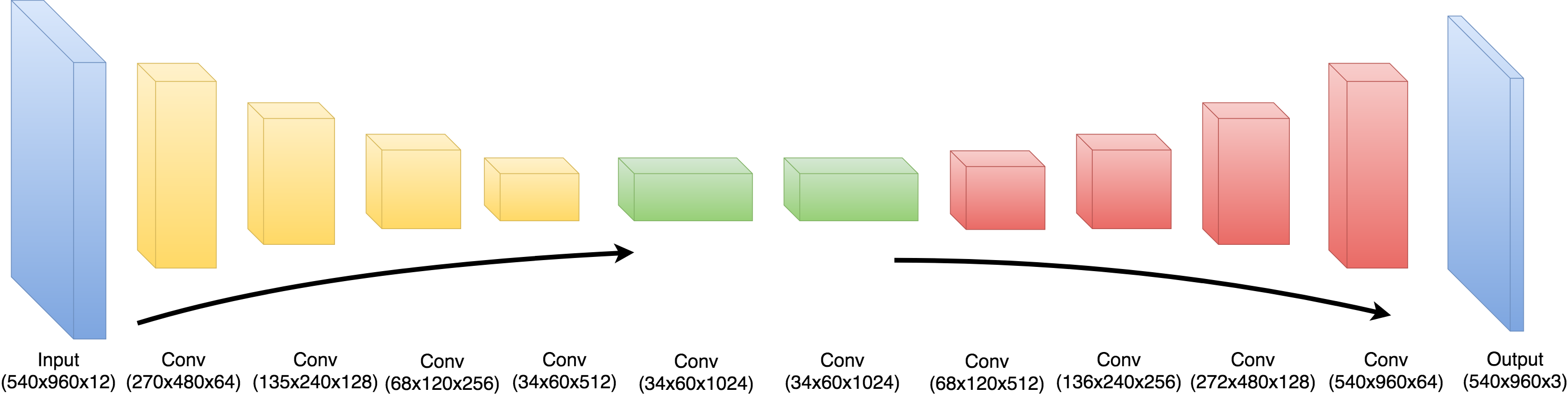}
      \caption{The proposed encoder-decoder CNN architecture. The contracting part is shown in yellow and the expanding part is in red.}
      \label{figure3}
  \end{figure*}

The proposed methodology for scene flow estimation uses an encoder-decoder architecture of convolutional neural network. Figure \ref{figure3} shows the network model with input, output and intermediate convolution layers. The network is fully convolutional. Thus, it can take input of any given size and produce spatial output. It consists of eleven layers. The dimension of the feature maps gradually decreases in the forward direction from input layer. This is achieved by applying strides while convolution operation. With every successive convolution layer the number of feature maps are doubled. This ensures that only most significant features are extracted. The expanding part of the CNN maps the feature maps to the output layer. The upconvolution in the expanding part is done by upsampling the input and subsequently applying convolution operation. To keep the output consistent with the dimension of scene flow ground truth, a small cropping function was applied after tenth layer. Table \ref{table1} shows the proposed network architecture in detail.
\begin{table}
\caption{Detailed Specifications of the SceneEDNet architecture.}
\label{table1}
\begin{center}
\resizebox{\linewidth}{!}{\begin{tabular}[!t]{|c|c c c|c c|c|}
\hline
Name & Stride & Ch I/O & In Res & Out Res & Input\\
\hline
conv0 & 3 & $12/64$  & $540\times960$ & $270\times480$ & Images\\
conv1 & 3 & $64/128$ & $270\times480$ & $135\times240$ & conv0 \\
conv1\_1&3  & $128/256$& $135\times240$ & $68\times120$  & conv1 \\
conv2 & 3 & $256/512$ & $68\times120$ & $34\times60$ &conv1\_1 \\
\hline
conv2\_1 & 3 & $512/1024$ & $34\times60$ & $34\times60$ & conv2\\
conv3 & 3 & $1024/1024$ & $34\times60$ & $34\times60$ & conv2\_1\\
\hline
conv3\_1 & 3 &$1024/512$ & $34\times60$ & $68\times120$ & conv3\\
conv4 & 3 &$512/256$ & $68\times120$ & $136\times240$ & conv3\_1\\
conv4\_1 & 3 & $256/128$ & $136\times240$ & $272\times480$ & conv4\\
conv5 & 3 & $128/64 $&$272\times480$ & $544\times960$ & conv4\_1\\
Output & 3 & $64/3$ & $540\times960$ & $540\times960$ & conv5\\
\hline
\end{tabular}}
\end{center}
\end{table}
  
The input layer is fed with stereo image pairs. The corresponding scene flow ground truth forms the output layer of the network. The kernel is a $3\times3$ filter which convolves over feature maps as it propagates in the forward direction while reduces the total number of trainable parameters. This kernel is used for all the convolution layers. For the problem formulation we tried two approaches. In the first approach, the target was a vector combining optical flow and depth at consecutive time steps with its corresponding image pairs as input. The network was very small with no change in dimensions. For the second approach, the output layer is sceneflow ground truth vector constructed from optical flow and disparity. The network in this case is of encoder-decoder form. For the second experiment the stereo images at consecutive time steps were concatenated. The target output is a combination of the depth and optical flow. The paper discusses details of the second approach. 
  
\section{Experiments and Results}
In this section we first provide a detailed description of the dataset and the annotation process. Next we discuss the implementation details of the proposed architecture, followed by the results obtained.
  \begin{figure*}[thpb]
      \centering
      \includegraphics[height=22cm, width=17cm]{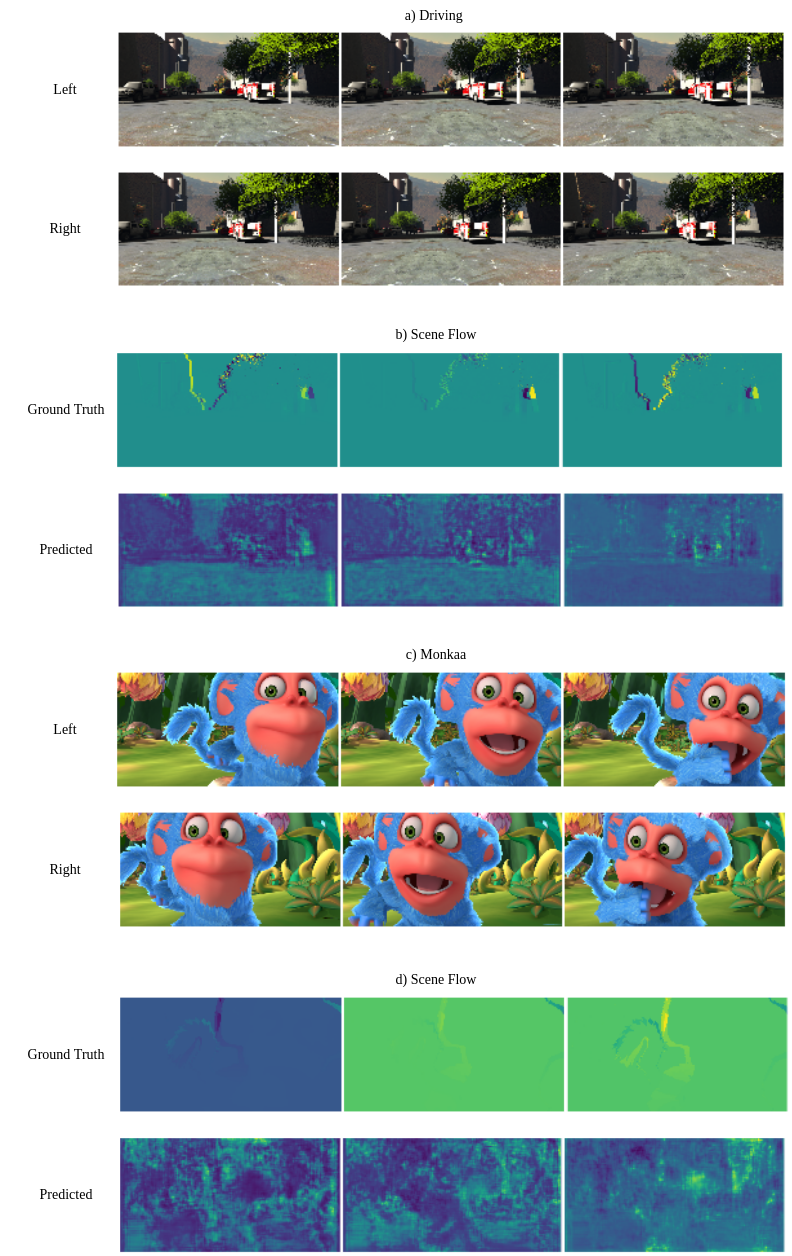}
      \caption{The result presented, are from first two images from each set. a) Driving stereo image pairs, The images from left to right images are at successive time steps b) Driving predicted and ground truth scene flow, From left to right, scene flow vectors in $x$, $y$ and $z$ direction. c) Monkaa stereo image pairs, From left to right image at successive time steps d) Monkaa predicted and ground truth scene flow. From left to right, scene flow vectors in $x$, $y$ and $z$ direction.}
      \label{figure4}
  \end{figure*}

\subsection{Dataset}
Training a convolutional neural network requires very large dataset. Many datasets like Middlebury \cite{scharstein2002taxonomy}, KITTI \cite{menze2015object} and Sintel \cite{Butler:ECCV:2012} are available with ground truth disparity and optical flow. However, they are not large enough to train a convolutional neural network. Thus, for direct estimation of scene flow we have trained our network using FlyingThings3D dataset \cite{mayer2016large}. This dataset is a part of scene flow dataset introduced by \cite{mayer2016large}. The dataset is artificially rendered in Blender using models from ShapeNet \cite{shapenet2015} database. The three dimensional models in the images also reflect changes in their position and orientation. The RGB images are available for both left and right view at different timesteps. The complete dataset is divided into three sets. There are different scenes in the dataset, with each scene containing fifteen frames. The RGB images are available in two versions. First, a clean pass, which is an ideal cases of input images. Second is a final pass version, where motion blur and other distortions are introduced. We have trained our network using final pass RGB images. The dataset also makes available the ground truth optical flow in both backward and forward directions. The disparity ground truth is in single channel image format. The dataset provides intrinsic and extrinsic camera parameters. For our work we have used forward optical flow and disparity to construct scene flow ground truth.

The scene flow dataset comes with RGB images for FlyingThings3D, Monkaa and Driving \cite{mayer2016large}. We used 18,000 stereo pairs from FlyingThings3D for training the network. Each frame has resolution of $540\times960\times3$. Though the ground truth optical flow and disparity are provided, we have constructed the scene flow.  The scene flow is considered as a 4D vector in the image space as  $(u,v,d_0, d_1)$ \cite{schuster2018combining}. Where $u$, $v$ are optical flow in horizontal and vertical direction respectively. Disparities are given by $d_0$ and $d_1$. This can be converted to world representation using camera parameters. To get the scene flow, we need to compute 3D position of each pixels in the consecutive time-steps. These positions can be obtained using stereo disparity relations and camera projection models. The scene flow is then obtained as translation vector from one 3D point to another. 

\subsection{Implementation Details}
The experimemt was performed using Keras deep learning framework \cite{chollet2015keras}. The proposed network was trained on all the three sets of FlyingThings3D. We used Stochastic Gradient Descent (SGD) as an optimizer. The SGD was observed to perform better than other popular optimizers like ADAM and RMS while experimentation. The learning rate was kept as 0.00001 with momentum at 0.5. For the activation function leaky relu was found to generate better output compared to the other activation functions. All the training parameters were found empirically. The one used for the training showed better convergence. The complete training was done for hundred epochs. We used a learning rate decay function to make sure that optimization process does not overshoot minimum value. This decay rate is a function of learning rate and total number of epochs. After every epoch the learning rate decreases according to the specified decay rate. Though this makes the training process longer, it makes sure that training loss and validation loss shows convergence. For the optimization process, a three dimensional end point error function was used as loss function. This end point error function is a euclidean distance between the predicted scene flow vector and ground truth scene flow vector. The scene flow network was trained on two Nvidia GTX-1080 GPUs. 

\subsection{Results}
The model was trained on FlyingThings3D \cite{mayer2016large} scene flow dataset. The trained model is used to obtain the results on sample images from Monkaa and Driving dataset. The test stereo pairs from both the subsets are cleanpass RGB images. Figure \ref{figure4} shows the predicted and ground truth scene flow with corresponding stereo images. The network is given stereo image pairs as input. The stereo images at $t$ and $t+1$ are concatenated to form a stack of input. The optical flow between $t$ and $t+1$ of left images is used along with two disparities to form scene flow ground truth vectors. With the forward pass over the trained network the scene flow vectors are obtained for $x$, $y$ and $z$ direction separately.t The total computation takes less than 10 seconds.

\section{CONCLUSION AND FUTURE WORK}
In this paper we presented an encoder decoder CNN architecture for scene flow estimation. We have shown that, with the availability of a large dataset with corresponding ground truth optical flow and disparity, training of a CNN for direct estimation of scene flow is possible. The proposed architecture cannot be claimed to perform better than the state-of-the-art scene flow estimation methods. However, it shows the way to apply an enhanced version of the proposed network, to find the scene flow efficiently. This deep learning method can be useful in cases where ideal assumptions of input images are violated. Moreover, the deep neural network trained in such cases has shorter run time than classical techniques. This makes it possible to deploy it for real time applications. The number of trainable parameters increases the network design. The network design can be made to reduce the number of parameters. Since, large number of parameters can limit the number of layers in neural network. Furthermore, the deep learning method for scene flow estimation can be improved by training it on naturalistic scenes.  

\section{ACKNOWLEDGMENTS}
The authors would like to thank Dr. Shivram Dubey and Soumen Ghosh, PhD Scholar, for their insightful discussions. 

\bibliographystyle{IEEEtran}
\bibliography{ref}






\end{document}